\pdfoutput=1
\documentclass[11pt]{article}
\usepackage{graphicx}

\usepackage{tikz}
\usepackage{amsmath} 
\usetikzlibrary{arrows.meta, positioning}

\usepackage[]{ACL2023}
\usepackage{graphicx}

\usepackage{times}
\usepackage{latexsym}

\usepackage[T1]{fontenc}
\usepackage[utf8]{inputenc}

\usepackage{microtype}

\usepackage{inconsolata}

\title{PedagoSense: A Pedology Grounded LLM System for Pedagogical Strategy Detection and Contextual Response Generation in Learning Dialogues}

\author{
Shahem Sultan\textsuperscript{1}
\And
Shahem Fadi\textsuperscript{2}
\And
Yousef Melhim\textsuperscript{2}
\And
Ibrahim Alsarraj\textsuperscript{3}
\And
Besher Hassan\textsuperscript{3}
\AND
\normalfont
\textsuperscript{1} Al Andlus University \\
\textsuperscript{2} Ajman University \\
\textsuperscript{3} Mohamed bin Zayed University of Artificial Intelligence \\
}

\begin{document}
\maketitle

\begin{abstract}
This paper provides a critical challenge to the enhancement of interaction quality in conversation-based learning environments through the detection and recommendation of effective pedagogical strategies employed by tutors. We propose a novel approach by incorporating NLP techniques on the modeling of classification of pedagogical strategies during tutor-student dialogues. Our approach involves a two-layer classifier. It first uses binary classification of pedagogical strategies and then performs fine-grained classification to classify the exact strategy used. Our experiments use conversation datasets thoroughly annotated with pedagogical strategy labels. Preliminary results show the effectiveness of our system with high accuracy in two kinds of classification tasks. Additionally, we utilized some latest models to recommend the most suitable pedagogical strategies in the context of the conversation. This study aspires to bridge the gap between pedagogical theory and its practical application in the development of more effective and personalized educational technologies. \href{https://github.com/IbrahimXXs/Tutor-Copilot-AI-Tutor-Response-Grounded-on-Pedology}{GitHub Repository}
\end{abstract}

\section{Introduction}
\subsection{Overview}

The project aims at the creation of Tutor Copilot system which would be capable of detecting pedagogical strategies used by tutors in the context of conversation-based learning as well as suggesting strategies. As the intelligent tutoring systems become more popular within the educational sphere, there is a need to simulate the interaction of tutors with their pupils in order to give them better individual pedagogical guidance. Finding and suggesting these strategies is common for this project and will help the development of gentle and smart tutoring systems to the students enhancing their experience wit. These insights will be instrumental in the development of more advanced and effective educational technologies in the area of natural language processing.

\subsection{Related Work}

Wang et al. \cite{wang2023} Investigates the involvement of LLMs to improve the performance of novice tutors with expert decisions within educational dialogues. they showed the feasibility of NLP systems to narrow down the knowledge disparity between the expert and the novice tutors through cognitive task analysis.

Demszky et al. \cite{demszky2022} In describing the work of tailor-made educational technologies. They addressed the problem of giving feedback to teachers and others that have employed the NLP systems to evaluate interactions in a classroom and concluded that using appropriate teaching methods could improve the attention of the learners to the lessons.

Li et al. \cite{li2020} Contemplated developing a conversational Ai for the purpose of evaluating and catagorizing tutor-student interactions in educational environment. There have been calls for more detailed coding of tutor’s dialogues to facilitate the comprehension of interactions in scholarly discussions.

Hosseini et al. \cite{hosseini2021} Presented a technique that employed deep learning models to reveal teaching strategies employed in e-learning environments. The accuracy focused more on inherent features of the specific strategies and placed the groundwork for further direction in strategy recognition using educational hyphen NLP strategies.

While building upon these topics and previous research, this project will strive towards incorporating pedagogical strategy detection within a unified NLP model. Specifically, we want to create a system, integrating foundational models and deep learning-based language models, to classify pedagogical strategies during two-way communication in real time. In this case, we are working towards enhancing intelligent tutoring systems in terms of providing more intelligible, personalized and context-sensitive pedagogical strategy suggestions in real time.

\section{Project Idea}

\subsection{Proposed Method}

The system we have developed makes it feasible for adaptive tutoring through the use of available data from the conversations between the student and the tutor to identify and utilize effective pedagogical strategies. pedagogical strategy could be: Probing, Giving a hint, provide a similar question or explain the concept etc. In this respect, for example, when a student communicates with the tutor asking that he is struggling with a particular problem or confusion, the fine-tuned BERT generating model looks into the conversations and situationally recommends a pedagogical strategy that seems to fit scenario well. After that, this recommended strategy goes to an LLM (GPT-4o through an API) which will generate a suitable response to be used by the tutor to explain the student.
As soon as an LLM produces a reply, it is subjected to another Binary-Classification model based on BERT Base which checks for the presence of any pedagogical response. In this context, “0” means no strategy has been used and “1” means a pedagogical strategy has been incorporated. When there is a strategy, it goes to Classifier fine-tuned BERT model in order to identify the type of the pedagogical strategy that has been used in the response. This classification is compared with the proposed one by the generating model; if there is a match, it confirms the proposed pedagogical strategy and verifies that our system is appropriate and useful. A video demonstrating our system Tutor Copilot can be viewed through the link \href{https://drive.google.com/file/d/1vAGxsKRggoC_gl4z0wKETvrj339NGzvH/view?usp=sharing}{Video}

\begin{figure}[h!]
  \centering
  \includegraphics[width=0.5\linewidth]{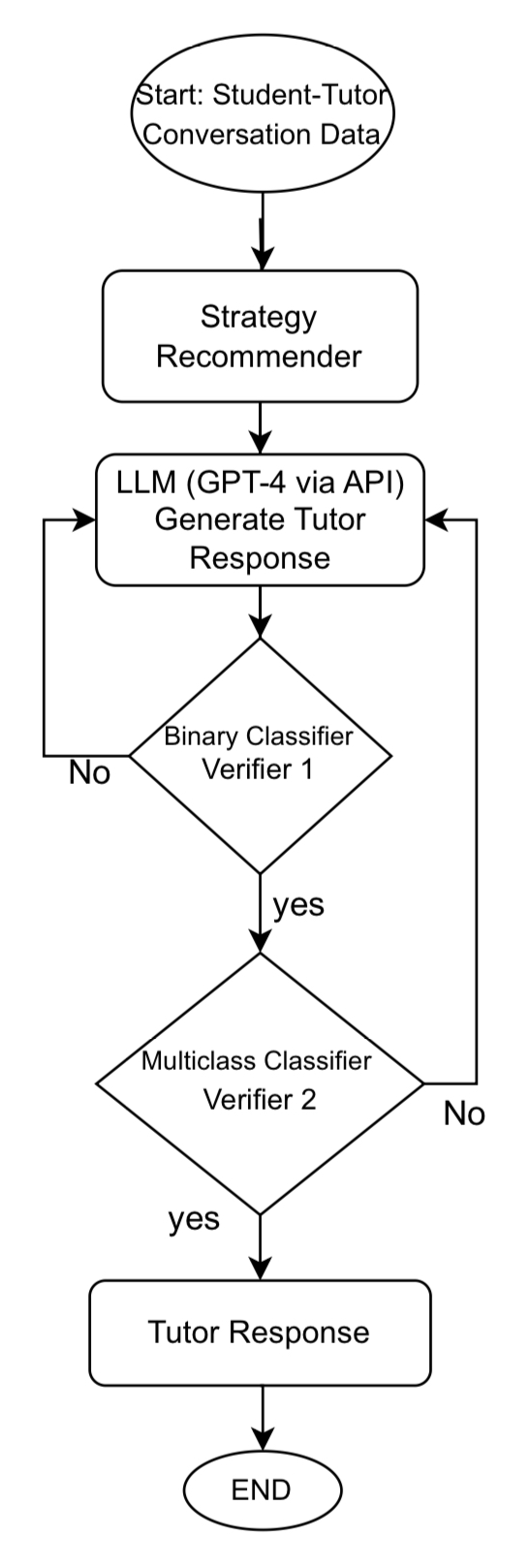} % The image file should be in the same directory as the LaTeX file or provide the path.
  \caption{Flow chart of the system}
  \label{fig:framework}
\end{figure}

\section{Dataset}

Two datasets are given by Dr. Kaushal Kumar Maurya and they include samples of the conversation between the tutor and student. As well as 697 samples from the Hugging Face DailyDialog dataset were used to augment the class 0 in the binary classification, and this dataset contain general conversations.

\subsection{Features for Binary Classification}

\noindent \textbf{Pedagogical Label}: In the Binary Classification task, a label of ‘0’ denotes that no strategy was employed whereas ‘1’ which indicates that a strategy was employed.

\noindent \textbf{Size}: 1.4K Pairs

\subsection{Features for Multi-Class Classification and Strategy Recommendation}

\noindent \textbf{Conversation History}: Every line in this data column includes a relatively semantic complete conversation between tutor and student, and this part is used as the input of the strategy prediction model.

\noindent \textbf{Expert Human Tutor Response}: The response from human tutor after the corresponding conversation history, and it's the input of the multi-class classification model.

\noindent \textbf{Pedagogical Strategy Label}: The details of the specific strategy applied by the tutor, such as Probing, Explain Concept, Provide Hint, or None. There are 8 distinct pedagogical strategies included in the Dataset after being pre-processed.

\section{Experiments and Evaluation}

\subsection{Binary Pedagogical Strategy Detection}

\subsubsection{Data Preparation and Preprocessing}
\begin{figure}[t]
    \centering
    \includegraphics[width=0.48\textwidth]{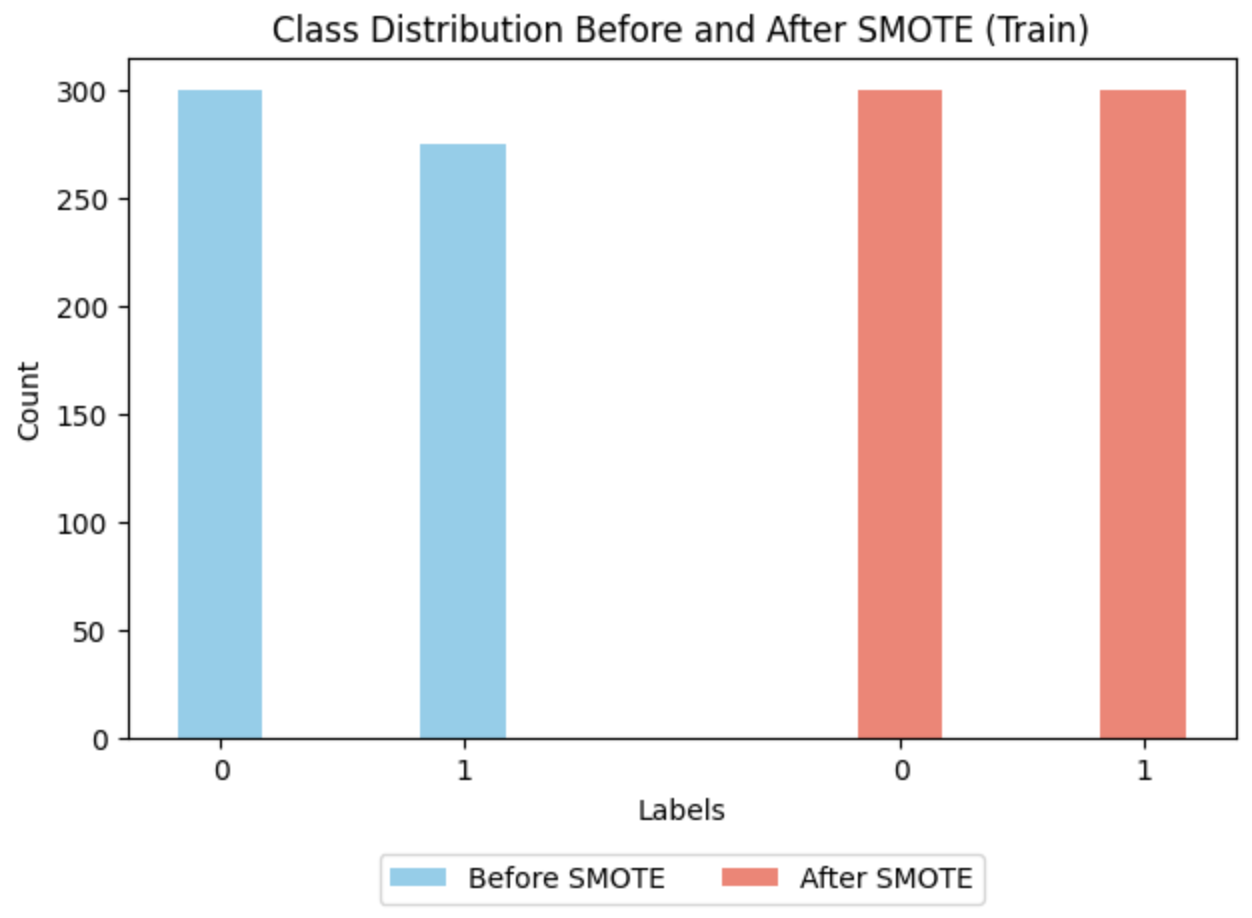}
    \caption{Class Distribution: Before and After SMOTE}
    \label{fig:smote_appendix}
\end{figure}

For this phase, I used the \texttt{Expert\_Human\_Tutor} feature, discarding the irrelevant features and removed duplicates to avoid contamination or bias. Preprocessing included lower casing, punctuation removal, stop word removal, and lemmatization, standardizing the data for better generalization. After removing  duplicates the dataset become imbalance with 303 samples for class 0 and 279 samples for class 1. So to balance the data the first approach used is SMOTE, which works by identifying each minority class sample's nearest neighbors in feature space, then creating new samples by interpolating between the original sample and its neighbors. This process generates realistic, diverse samples, and apply balance without simply duplicating data. Therefore, it increased the samples to 303 for each class as shown in Figure~\ref{fig:smote_appendix}. While the SMOTE is an easy and fast solution to balance the data, techniques like data augmentation outperform it by making the quality of the generated data better, which improves the performance. So for the  the second approach data augmentation have been applied by using GPT-4o API. For class 1, I generated 721 additional samples using the GPT-4o API, increasing the total to 1,000 samples. This focused on examples containing pedagogical strategies. For class 0, I incorporated 697 samples from the Hugging Face DailyDialog \cite{li2017dailydialog} dataset, which includes general-purpose conversations without pedagogical strategies. This increased the total to 1,000 samples. The final training dataset was balanced, with 2,000 samples equally distributed across both classes. And the testing and validation dataset remain untouched with 223 samples for the test set and 73 samples for the validation set.

While SMOTE effectively balanced the dataset, it introduced noise in high-dimensional data, leading to potential overfitting. This highlighted the need for more robust methods, such as data augmentation, to enhance generalization.

\subsubsection{Model Selection and Setup}
I selected Bert Base, for binary classification tasks. And I used the Hugging Face library to fine-tune the model with tokenized text data, setting a maximum token length of 128 to ensure consistency across the dataset. Key training parameters included a learning rate of $5 \times 10^{-5}$, a batch size of 16, and an early stopping mechanism to mitigate overfitting. Early stopping optimized training performance and generalization by halting training when validation loss ceased to improve.

\subsubsection{Baseline and Evaluation Metrics}
The baseline model was a Logistic Regression model utilizing TF-IDF features, achieving F1-scores of 95.83\% on the validation set and 94.90\% on the test set. Both the baseline and BERT models were evaluated using accuracy, precision, recall, F1-score, and loss metrics, with specific BERT performance details provided in the Experimental Results section.

\subsubsection{Experimental Results}
The BERT model with data augmentation using GPT-4o, achieved its best validation F1 score of 98.85\% at the 6th epoch, with a validation loss of 0.0636 over 10 epochs. On the test set, it recorded an F1 score of 98.5\%, showing a significant improvement over the baseline Logistic Regression model. Table \ref{tab:model_comparison} compares F1 scores across models for binary pedagogical strategy detection, incorporating SMOTE and Data Augmentation. The SMOTE Improved performance for Naive Bayes and SVM models. While Data Augmentation Increased F1 scores for most models, particularly fine-tuned BERT, MLP, and LightGBM. Random Forest and Decision Tree Performance declined with Data Augmentation, likely due to sensitivity to noise or overfitting. BERT Base consistently delivered the highest scores, confirming its effectiveness in NLP tasks within this use case.

\begin{table}[ht]
\centering
\resizebox{0.48\textwidth}{!}{%
\begin{tabular}{|l|c|c|}
\hline
\textbf{Model} & \textbf{F1-Score (SMOTE)} & \textbf{F1-Score (Data Augmentation)} \\ \hline
Naive Bayes & 97.73 & 98.0 \\ \hline
BERT Base After Fine-Tuning & 97.34 & \textbf{98.5}\\ \hline
SVM & 97.34 & 96.0 \\ \hline
MLP & 96.55 & 98.0 \\ \hline
Logistic Regression & 95.31 & 95.0 \\ \hline
Random Forest & 93.02 & 87.0 \\ \hline
Decision Tree & 88.16 & 80.0 \\ \hline
XGBoost & 86.61 & 92.0 \\ \hline
AdaBoost & 86.07 & 85.0 \\ \hline
LightGBM & 84.13 & 90.0 \\ \hline
LDA & 79.38 & 95.0 \\ \hline
BERT Before Fine-Tuning & 77.0 & 77.0 \\ \hline
KNN & 22.82 & 52.0 \\ \hline
\end{tabular}%
}
\caption{F1-Score Comparison: SMOTE vs Data Augmentation on test set}
\label{tab:model_comparison}
\end{table}

\subsubsection{LLMs Evaluation}
\begin{figure}[!h]
    \centering
    \includegraphics[width=0.48\textwidth]{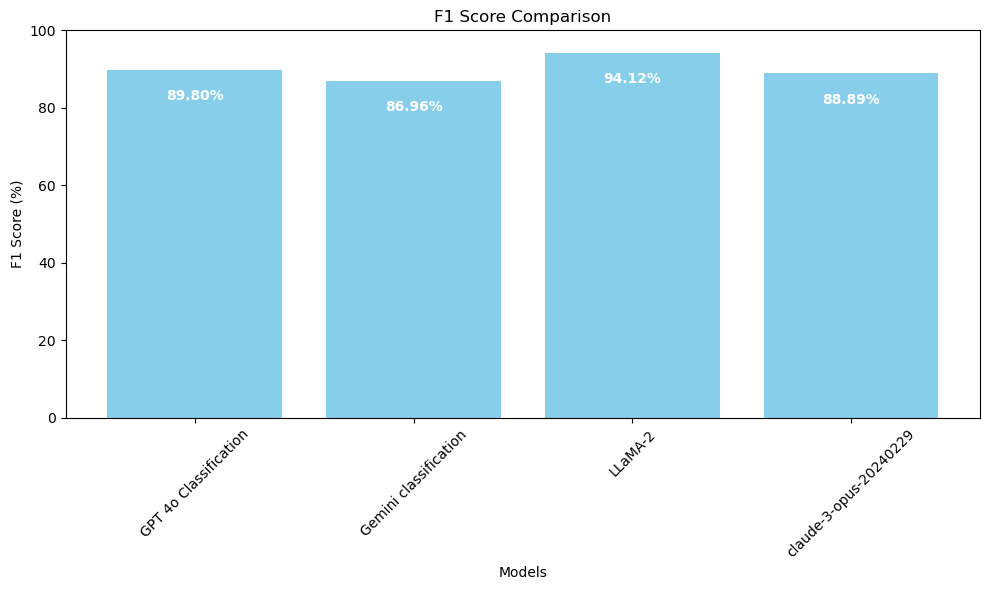} % Adjusted size
    \caption{Word Contributions to Classification}
    \label{fig:llms}
\end{figure}

Figure \ref{fig:llms} presents the F1-scores of four large language models (LLMs) evaluated on 50 samples, 25 of the from class 0 and the other 25 from class 1 to check  their ability to identify pedagogical strategies. LLaMA-2 Scored the highest at 94.12\%, showing superior performance in analyzing and identifying educational strategies. And GPT-4o Achieved 89.80\%, performing well but slightly below LLaMA-2. while Gemini and Claude Scored significantly lower, showing a challenges in detecting the pedagogical strategies.

\subsubsection{Analysis}
\begin{figure}[!h]
    \centering
    \includegraphics[width=0.48\textwidth]{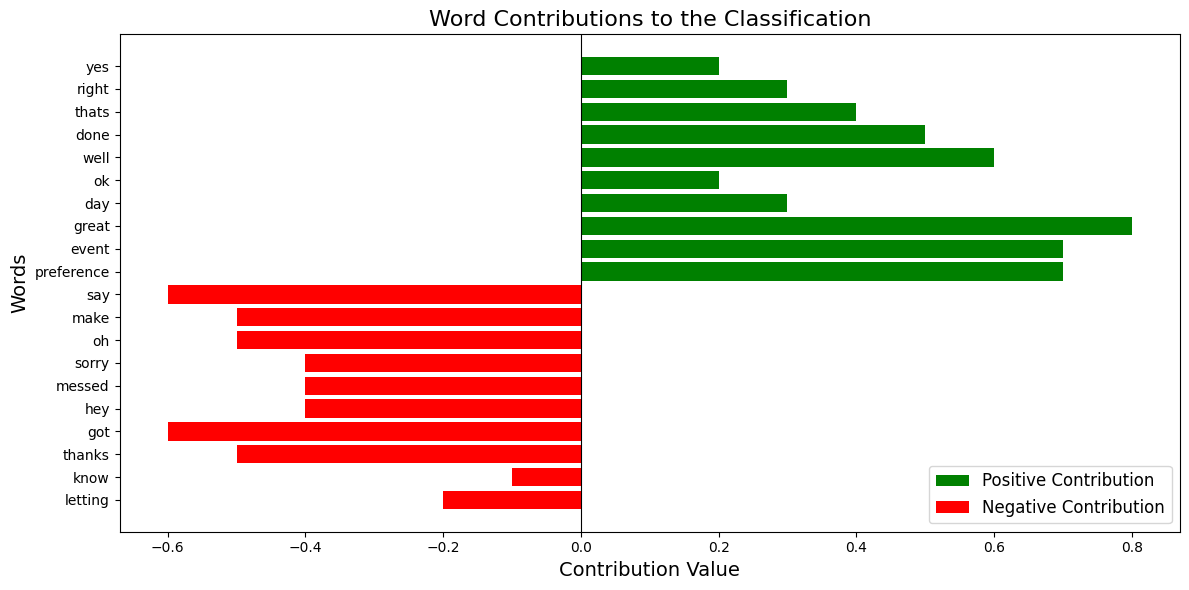} % Adjusted size
    \caption{Word Contributions to Classification}
    \label{fig:lime}
\end{figure}

To explain the BERT model's predictions, I applied LIME (Local Interpretable Model-agnostic Explanation) to the five misclassified samples, highlighting words that were critical to each prediction. As shown in Figure ~\ref{fig:lime}, LIME makes it possible to observe how certain words contribute positively or negatively to the model's decisions. For instance, words like “great” had a high positive impact, which leads to misclassifications and increasing the likelihood of wrong class predictions, while words like “say” and “got” contributed negatively, which shifted predictions away from the actual target class. These observations suggest that the model may be over-relying on a few prominent keywords, rather than considering the broader context tendency that is not optimal for nuanced predictions. Furthermore, the impact of SMOTE and Data Augmentation on model performance highlights the different patterns. For instance, Data Augmentation improved the performance for models like BERT and MLP by introducing diverse training samples, enabling these models to better generalize. However, it produced noise for tree-based models like Random Forest, which may have decreased their ability to differentiate meaningful patterns, highlighting the importance of tailoring preprocessing techniques to the characteristics of each model type.

\subsection{Fine-Grained Classification}
\subsubsection{Data Preparation and Preprocessing}\label{data_clean}

Due to the small dataset size, quality is crucial. I removed all labels with fewer than 20 occurrences, as the model cannot effectively learn from them, and they act as noise, impairing performance. The two primary columns that I used are (Expert\_Human\_Tutor) which is the tutor's reply to the student after a conversation, and (Peda\_Strategies) which is the target pedagogical strategies. The input was tokenized using BERT tokenizer which converted the text into IDs with a maximum sequence of 128 tokens. If the input exceeds the token limit, it will get truncated. On the other hand, padding tokens were added to shorter sequences. Attention masks were implemented as well to differentiate between real tokens and padding tokens. Attention masks and padded input sequences were converted into tensors to be used as the input for the model. The target pedagogical strategies got encoded using a Label-Encoder which transforms string labels into number values suitable for training.

\subsubsection{Data Augmentation}

\begin{figure}[h!]
  \centering
  \includegraphics[width=0.98\linewidth]{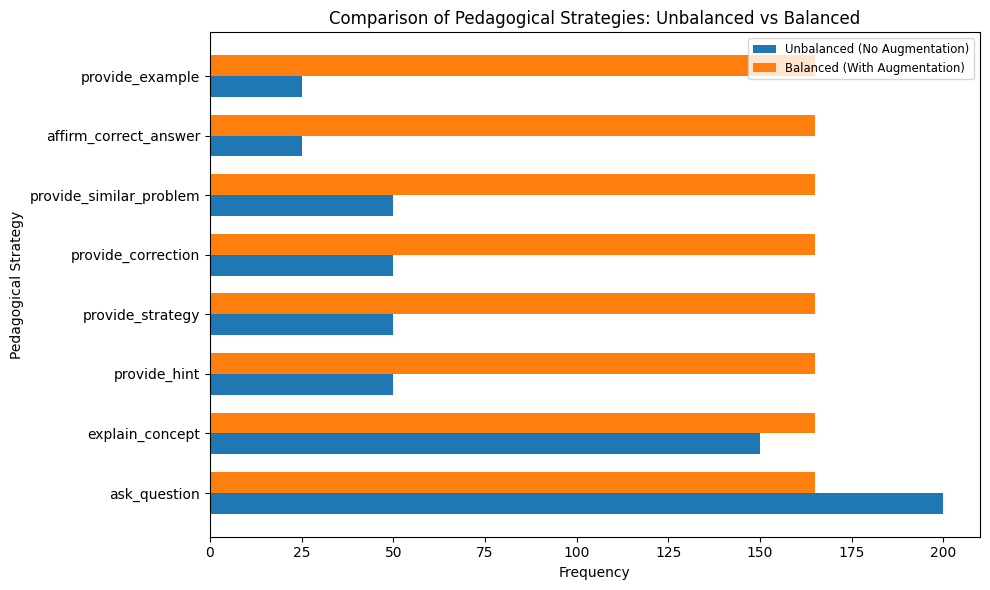}
  \caption{Comparison of Pedagogical Strategies: Unbalanced vs Balanced.}
  \label{fig:freq}
\end{figure}

Given the small size and the unbalanced distribution of pedagogical classes in the available dataset. GPT-4o was utilized to generate and augment Expert\_Human\_Tutor with the target pedagogical strategy in order to have a more generalized model. Figure \ref{fig:freq} shows the frequency of each pedagogical strategy before and after data augmentation.

\subsubsection{Model Selection and Setup}

I chose BERT Large uncased pre-trained transformer language model for fine-tuning. The uncased version was used because it doesn't differentiate between uppercase and lowercase which makes it a robust model. AdamW optimizer with weight decay was used to update the model weights during training. A variant which is designed specifically for classification tasks BertForSequenceClassification was employed from the hugging face transformers library where the output which passes through SoftMax layer will have a probability distribution over all eight different pedagogical strategies. Table \ref{tab:hyperparameters} shows the best selected hyperparameters after testing.
\begin{table}[ht]
    \centering
    \resizebox{0.48\textwidth}{!}{ % Resize the table to 50% of the text width
    \begin{tabular}{l|c}
    \hline
    \textbf{Hyperparameter} & \textbf{Value} \\
    \hline
    Learning Rate           & $2 \times 10^{-5}$ \\
    Batch Size              & 16 \\
    Epochs                  & 20 (with early stopping) \\
    Optimizer               & AdamW (with weight decay) \\
    Max Sequence Length     & 128 tokens \\
    \hline
    \end{tabular}
    }
    \caption{Fine-Grained Classification Model hyperparameters.}
    \label{tab:hyperparameters}
\end{table}

\subsubsection{Baseline and Evaluation Metrics}

My baseline will be the smaller version of BERT uncased which is BERT base uncased model.

\subsubsection{Experimental Results And Analysis}\label{ER}

\begin{table}[ht]
    \centering
    \resizebox{0.5\textwidth}{!}{ % Resize to 50% of the text width
    \begin{tabular}{l|c|c}
    \hline
    \textbf{Metric}             & \textbf{BERT Large Uncased (\%)} & \textbf{BERT Base Uncased (\%)} \\
    \hline
    Validation Accuracy         & 60.27                            & 57.53                           \\
    Test Accuracy          & 56.65                            & 53.76                           \\
    Macro F1 Score         & 45.95                            & 44.87                           \\
    \hline
    \end{tabular}
    }
    \caption{Models Performance Metrics (Augmented Balanced Dataset}
    \label{tab:model_performance_transposed_resizedz}
\end{table}

Table \ref{tab:model_performance_transposed_resizedz} Shows the validation accuracy, test macro F1 score and test accuracy between two different BERT variants on the augmented balanced dataset.

\begin{table}[ht]
    \centering
    \resizebox{0.5\textwidth}{!}{ % Resize to 50% of the text width
    \begin{tabular}{l|c|c}
    \hline
    \textbf{Metric}             & \textbf{BERT Large Uncased (\%)} & \textbf{BERT Base Uncased (\%)} \\
    \hline
    Validation Accuracy         & 55.15                            & 57.53                           \\
    Test Accuracy               & 49.13                            & 49.13                           \\
    Macro F1-Score              & 42.17                            & 32.67                           \\
    \hline
    \end{tabular}
    }
    \caption{Models Performance Metrics (Unbalanced Dataset | With No Data Augmentation}
    \label{tab:model_performance_transposed_resized}
\end{table}

On the other hand, Table \ref{tab:model_performance_transposed_resized} Shows the validation accuracy, test macro F1 score and test accuracy between two different BERT variants on the unbalanced dataset set with data augmentation.

Table \ref{tab:class_accuracyz} demonstrate the average class accuracy for each pedagogical strategy class for BERT Large Uncased on the augmented balanced dataset. More experimental results on different models can be found at the appendix \ref{extra}

\begin{table}[ht]
    \centering
    \resizebox{0.47\textwidth}{!}{ % Resize to 50% of the text width
    \begin{tabular}{l|c}
    \hline
    \textbf{Class}                & \textbf{Average Accuracy (\%)} \\
    \hline
    affirm\_correct\_answer       & 66.67                          \\
    ask\_question                 & 88.17                          \\
    explain\_concept              & 67.86                          \\
    provide\_correction           & 28.94                          \\
    provide\_example              & 25.25                          \\
    provide\_hint                 & 26.67                          \\
    provide\_similar\_problem     & 30.00                          \\
    provide\_strategy             & 26.60                          \\
    \hline
    \end{tabular}
    }
    \caption{Pedagogical Classes Average Accuracy.}
    \label{tab:class_accuracyz}
\end{table}

Based on the experimental results, 4 observations can be made:

1) Both pre-trained BERT variants showed an improvement in performance in terms of validation accuracy, test accuracy and macro F1 score. For instance, the macro F1 score BERT Large Uncased improved from 42.17\% to 45.95\% demonstrating the effectiveness of balancing the unbalanced pedagogical strategies.

2) BERT Large Uncased showed a consistently higher performance than BERT Base Uncased. For example, BERT Large achieved a macro F1 score of 45.95\% compared to 44.87\% for BERT Base showcasing the ability of detecting and capturing nuanced patterns in larger models, though the margin is not substantial.

3) Pedagogical strategies like provide\_example and provide\_hint stand as major problem to the model making it struggles to differentiate. Compared to high average accuracy classes like ask\_question and explain\_concept, suggests that they have more subtle or ambiguous distinctions which requires more data augmentation and robust representation.

4) High accuracy dominant pedagogical classes heavily contributes to the overall macro F1 score.

\subsection{Pedagogical Strategy Recommendation}

\subsubsection{Dataset Cleaning}

% # 数据集清洗
%标签清洗、标签合并、分隔符添加
%由于数据集的规模很小，因此质量至关重要。我删除了所有出现频率小于20条的标签，显然模型无法学到该标签，反而这类标签会作为噪声影响模型效果；最终我保留的教学策略标签包括[affirm_correct_answer, ask_question,encourage_student,explain_concept,provide_correction,provide_hint,provide_similar_problem,provide_strategy,simplify_question].
Same as Section \ref{data_clean}, only labels with a frequency greater than 20 in the initial dataset were retained. In this task, the only feature data column used was Conversation History.

\subsubsection{Baseline Models}\label{baseline}

\textbf{Hybrid-Traditional Voting.} This method leverages the complementary strengths of three traditional machine learning models — Support Vector Machine (SVM), Naive Bayes, and Boosting — to predict pedagogical strategies through a simple majority voting mechanism. Each model was independently trained on the same dataset, which was preprocessed using stopword removal, stemming, label encoding, and transformation of text features into TF-IDF vectors to ensure consistent feature representation. SVM, known for its ability to model high-dimensional feature spaces with separating hyperplanes, excelled in precision. Naive Bayes, a probabilistic classifier assuming feature independence, demonstrated competitive recall, while Boosting, which sequentially improves classification by focusing on harder-to-classify instances, achieved balanced overall performance. By aggregating the predictions from these individual models through majority voting, the Hybrid-Traditional Voting method effectively combines their strengths, resulting in improved recommendation accuracy for pedagogical strategies. Metrics for this baseline model can be found in Table \ref{tab:metrics}.

\noindent \textbf{Recommendation Based on Label Probability Distribution (LPD).} Assuming that the current dataset represents the optimal real-world adoption of teaching strategies, the frequency of different strategy labels in the dataset inherently reflects knowledge about their usage by human teachers. For instance, if the strategy 'ask question' appears three times more frequently than other strategies, it is reasonable to recommend it in most cases. Therefore, in the baseline recommendation model, I used the label frequency distribution as probabilities for suggesting teaching strategies. Metrics for this baseline model can be found in Table \ref{tab:metrics}.

%以承认当前数据集就是最佳现实世界教学策略采用方案为前提，可知数据集中不同策略标签被人类教师使用的频率本身就已经包括了某种关于策略使用的知识。比如教学策略‘ask question’的出现次数超过其他策略3倍，那么大多数情况下可以直接推荐教师使用这个策略。因此，在最基础的推荐模型中我直接根据标签的出现频率分布作为概率来推荐教学策略。该基线模型的各项metric可以在表1中查看。
\noindent \textbf{Recommendation Combining BM25 and Embedding Similarity (BES).} I selected the top-$k$ ($k=5$) most similar conversation histories based on the BM25 score of the current conversation history. I then calculated the embedding similarity between the current conversation history and these $k$ similar conversations. I introduced a hyperparameter $\alpha$ to balance the two methods, with the final score calculated using the formula: $(\alpha \times \text{bm25\_score} + (1 - \alpha) \times \text{embedding\_scores}) \times \text{strategy\_prob}$, where \text{strategy\_prob} denotes the prior probability of the label. The model achieved the best result when $\alpha=0.2$. This method recommends the pedagogical strategy with the highest combined score. Metrics for this baseline model are shown in Table \ref{tab:metrics}.

\noindent \textbf{BERT-Base.} This method utilizes the pre-trained BERT model, specifically \texttt{bert-base-uncased}, without any data augmentation, to predict pedagogical strategies. The model is fine-tuned on the cleaned dataset using a batch size of 8, the AdamW optimizer with a learning rate of $2 \times 10^{-5}$, and trained for 20 epochs. BERT-Base leverages its pre-trained language representation to classify strategies directly based on the input conversation history. However, this method performs poorly, even slightly worse than the \textit{Hybrid-Traditional Voting} method, highlighting the challenges posed by the limited dataset size. Metrics for this baseline model are summarized in Table~\ref{tab:metrics}.

\subsubsection{Data Augmentation}

I reviewed all 677 entries in the dataset and discovered two key findings. 1) The dataset size is clearly too small to enable the model to learn sufficient knowledge for effective strategy recommendation. 2) There is significant overlap and confusion between different labels. For instance, conversation histories labeled as requiring the "provide strategy" approach were not distinctly different from those needing the "provide hint" approach; in essence, either label could be reasonably applied to these histories. 

To enrich the dataset's information, I used the GPT-4o-2024-08-06 model for data augmentation. Specifically, I generated: 1) Explanations for teaching strategies, explaining the rationale for why the current conversation history should use the corresponding pedagogical strategy; 2) Lists of specific keywords or phrases as 'clues' to help identify similar contexts where current strategy would be effective. These were added as two new columns in the dataset. 

Notably, to ensure structured output from the API with formatted strings and lists, I utilized OpenAI's latest ‘Structured Outputs’ technology, which required predefining return classes in the code.

%我检查了数据集中所有700条数据，发现了2个事实。1.显然数据数量太少，不足以支撑模型学到足够的知识进行策略推荐；2.不同标签之间有明显的交叉混淆。比如，某些应当推荐provide strategy策略的对话历史和某些应当推荐provide hint策略的对话历史并没有模式上的区别，本质来说，这些对话历史属于这两个标签的任何一个都是合理的。因此，为了增加数据集中的知识，我调用gpt-4o-2024-08-06进行了数据增强。具体而言，我生成了1.对于教学策略的解释，explain the rationale for using this teaching approach；2.对于教学策略的线索，a list of specific keywords or phrases as 'clues' that could help identify similar contexts where this approach would be effective.，并将它们作为新的两列加入数据集。值得一提的是，为了控制api返回规范的字符串和列表，我使用了OpenAI API最新提供的Structured outputs，需要在代码中预定义返回class形式。另外，我对新的两个特征做了消融实验，对比结果可以在table1中找到。

% 为了确保结构化返回，使用了最新的structure；prompt设计
% 展示一个例子，explanation长什么样子，clue长什么样子

% #T5模型和bert-uncased模型
\subsubsection{Proposed Models}
% 简单介绍模型，然后简单讲代码流程
% \textbf{Recommendation Based on T5 Model}. T5 (Text-To-Text Transfer Transformer) is a generative model by Google that treats all NLP tasks as text-to-text, excelling in tasks like translation and summarization. Specifically, I used the \texttt{transformers} library to load the T5 model and tokenizer, created embeddings for each unique pedagogical strategy label, and updated model parameters using the AdamW optimizer (lr=5e-5). I trained the model with a batch size of 4 for 10 epochs. Then I used the model to generate outputs on the test set. I wrote a custom \texttt{get\_closest\_label} function to compute the cosine similarity between the generated results and the predefined label embeddings to identify the closest label. Metrics for this proposed model can be found in Table \ref{tab:metrics}.

\noindent \textbf{BERT-Augmented.} I used \texttt{LabelEncoder} to convert strategy labels from strings to numeric values for training. For data augmentation, I utilized the OpenAI GPT API to generate strategy explanations and clue lists for each data instance. These augmented texts were concatenated to the conversation history, forming enriched input data. I loaded the \texttt{bert-base-uncased} model and corresponding tokenizer from the \texttt{transformers} library and configured the output layer for classification to match the number of unique strategy labels in the dataset. \texttt{TeachingStrategiesDataset}, which inherits from \texttt{torch.utils.data.Dataset}, tokenizes and encodes the augmented input text and labels, returning PyTorch tensors for \texttt{input\_ids}, \texttt{attention\_mask}, and \texttt{labels}. I trained the model using a batch size of 8 for 10 epochs and updated the model parameters with the AdamW optimizer (\texttt{lr=2e-5}). I selected the predicted label from the logits using \texttt{torch.argmax}. Metrics for this augmented model can be found in Table \ref{tab:metrics}.

\noindent \textbf{Hybrid-BERT Voting Method.} The Hybrid-BERT Voting method integrates predictions from four sources: BM25, embedding similarity, label probabilities, and the data-augmented BERT model. Each source independently predicts pedagogical strategies for a given input. Through experimental observations, I found that these methods often make correct predictions on different instances. Motivated by this insight, I sought to combine these methods to fully exploit the potential of the dataset. The final recommendation is determined through a simple majority voting mechanism, aggregating the predictions from these diverse methods. By leveraging both traditional information retrieval techniques and the strength of the data-augmented BERT model, this method aims to improve the robustness and accuracy of pedagogical strategy recommendations. Metrics for this method are provided in Table \ref{tab:metrics}.

\noindent \textbf{Hybrid-BERT Probabilistic Voting Method.} Similar to the Hybrid-BERT Voting method, the Hybrid-BERT Probabilistic Voting method integrates predictions from four sources. However, instead of relying on a simple majority voting mechanism to select a single strategy, this method adopts a probabilistic approach. I computed the likelihood of each pedagogical strategy by combining probabilities from all sources using a weighted sum. The weights for the combination are as follows: \texttt{BERT} (0.5), \texttt{LPD} (0.2), and \texttt{BES} (0.3). This approach generates a ranked list of recommendations based on the final computed probabilities, providing a nuanced representation of the model's confidence in each strategy. Unlike the Hybrid-BERT Voting method, this probabilistic approach offers a more flexible framework for applications requiring multiple suggestions or prioritization of strategies. Metrics for this method are shown in Table \ref{tab:metrics}.

% #消融实验及实验结果分析
\begin{table}[ht]
    \centering
    \small  % or \footnotesize, \scriptsize
    \begin{tabular}{l|ccc}
        \hline
        \textbf{Metric} & \textbf{Precision} & \textbf{Recall} & \textbf{F1 Score} \\
        \hline
        \textbf{Hybrid-Trad} & 0.2543 & 0.3212 & 0.2306 \\
        \textbf{LPD} & 0.2118 & 0.2424 & 0.2247 \\
        \textbf{BES} & 0.1562 & 0.3182 & 0.2026 \\
        \textbf{BERT-Base} & 0.1909 & 0.3109 & 0.2000 \\
        \textbf{BERT-Augmented} & 0.3717 & 0.3729 & 0.4663 \\
        \textbf{Hybrid-BERT} & 0.5470 & 0.5389 & \textbf{0.4752} \\
        \textbf{Hybrid-BERT Prob} & 0.5546 & 0.5337 & \textbf{0.4815} \\
        \hline
    \end{tabular}
    \caption{Performance metrics on the test set.}
    \label{tab:metrics}
\end{table}

% \begin{table}[ht]
%     \centering
%     \begin{tabular}{l|ccc}
%     \hline
%           & \textbf{F1} & \textbf{Recall} & \textbf{Precision} \\
%     \hline
%     LPD   & 0.2247 & 0.2424 & 0.2118\\
%     BES   & 0.2026 & 0.3182 & 0.1562\\
%     T5    & 0.1661 & 0.2073 & 0.1497\\
%     T5+EX & 0.4071 & 0.4663 & 0.3717\\
%     T5+CL & 0.3620 & 0.4560 & 0.3729\\
%     T5+EX+CL & 0.4223 & 0.4974 & 0.3824 \\
%     BERT  & 0.2000 & 0.3109 & 0.3109\\    
%     BERT+EX & 0.7190 & 0.7358 & 0.7298\\
%     BERT+CL & 0.7023 & 0.7150 & 0.7257\\
%     BERT+EX+CL & 0.7081 & 0.7254 & 0.7090 \\
%     \hline
%     \end{tabular}
%     \caption{Performance metrics on the test set.}
%     \label{tab:metrics}
% \end{table}

%根据实验结果，可以观察到以下几点事实。1. 由于LPD方法的三项分数优于数据增强之前的T5方法，因此我们可以认为策略标签本身的概率分布确实如baseline小节中所分析的具有 inherently reflects knowledge about their usage by human teachers. 2.BES方法中当alpha=0.2时获得最优结果，说明Embedding Similarity方法占主导作用，比BM25算法更有效。3.BERT方法的效果整体显著优秀于T5方法，原因在于T5是生成模型，我们后置了计算输出和策略标签嵌入的相似度来获取标签，然而BERT方法中将分类层内置于网络结构，参数在训练过程中得到优化，因此分类效果强于T5. 4. 关于数据增强特征的消融实验，可以明显观察到EX特征比CL特征更有效。另外，在T5方法中，同时使用两种增强特征会得到更好的结果，但在BERT方法中仅使用增强特征EX会得到更好的结果，也就是说CL特征反倒使BERT的结果下降，这是有待讨论的。 5.可以明显观察到，无论是T5方法还是BERT方法，都在数据增强之后获得了显著提升，也就是说模型和方法本身是没有问题的，最大的问题在于数据集本身的信息信息量，我的未来计划也将由此入手，具体参见5.3小节。
% 上表（包括基线模型的结果和消融实验的结果），然后分析数据集有多烂，说clue效果没explanation好，但是两个组件都有用
%1. 概率信息有用；2.因为最佳a=0.2所以嵌入相似度比bm25更有用；3. bert整体比t5好是因为bert是分类模型t5是生成模型；4.原始模型比数据增强之后效果差很多，说明模型没问题，方法没问题，数据集问题很大，说明机器学习的关键是数据集的质量。更详细的改进计划在第5.3节。

Based on the experimental results, the following 5 observations can be made:

1) The LPD method outperformed the BERT-Base method in F1 Score, confirming that the probability distribution of strategy labels inherently reflects knowledge about their usage by human teachers, as discussed in the \ref{baseline}. 

2) The BES method achieved the best results when $\alpha$=0.2, indicating that the Embedding Similarity approach played a dominant role and was more effective than the BM25 algorithm. 

3) The Hybrid-BERT Probabilistic Voting method achieved the best results with the weights set as follows: \texttt{BERT} (0.5), \texttt{LPD} (0.2), and \texttt{BES} (0.3). This configuration indicates that the data-augmented BERT model played a dominant role in the combined predictions, providing highly effective probabilities derived from enriched input data. The moderate contribution of the \texttt{BES} component highlights the importance of leveraging contextual relevance through embedding similarity and BM25 scores, while the smaller weight for \texttt{LPD} reflects its role as a supplementary prior, ensuring alignment with label frequency distributions.

4) Significant improvements in performance were observed after data augmentation, highlighting the critical role of data volume in machine learning tasks. Since a model’s knowledge is entirely derived from the data it learns from, data availability often becomes the primary bottleneck for improving model performance.

5) Our observations revealed that different algorithms excelled at making correct predictions on varying instances, highlighting their complementary strengths. To fully leverage the dataset's potential, we implemented both simple majority voting and probabilistic voting strategies to integrate the outputs of all algorithms. This combination approach effectively capitalized on their individual strengths, resulting in improved recommendation performance and a more robust strategy prediction system.

% #未来计划
% 写在他俩标的地方

\begin{table}[b]
    \centering
    \resizebox{0.98\textwidth}{!}{ % Resize to 50% of the text width
    \begin{tabular}{l|c|c|c}
    \hline
    \textbf{Model}                      & \textbf{Validation Accuracy (\%)} & \textbf{Test Macro F1 Score (\%)} & \textbf{Test Accuracy (\%)} \\
    \hline
    BERT Large Uncased                  & 60.27                            & 45.95                            & 56.65                         \\
    BERT Base Uncased                   & 57.53                            & 44.87                            & 53.76                         \\
    Multi-Layer Perceptron              & 51.22                            & 44.70                            & 50.41                         \\
    Naive Bayes                          & 52.03                            & 40.43                            & 46.34                         \\
    Support Vector Machines (SVM)        & 56.10                            & 41.17                            & 45.53                         \\
    Random Forest                        & 47.15                            & 36.97                            & 45.53                         \\
    Logistic Regression                  & 56.10                            & 40.29                            & 44.72                         \\
    RoBERTa Base                         & 49.45                            & 40.37                            & 44.51                         \\
    k-Nearest Neighbors                  & 47.15                            & 35.50                            & 41.46                         \\
    Decision Trees                       & 40.65                            & 32.99                            & 40.65                         \\
    Perceptron                           & 53.66                            & 32.94                            & 38.21                         \\
    \hline
    \end{tabular}
    }
    \caption{Performance on Different Models.}
    \label{tab:model_performancex_resized}
\end{table}

\section{Conclusion and Future Work}\label{conclusion}

In conclusion, we created Tutor Copilot, which is a system for classifying and recommending pedagogical strategies in dialogue-based learning with a novel approach that combines a two classification layers: binary classification, which detects the presence of the pedagogical strategy, and fine-grained classification to classify the type of the pedagogical strategy alongside strategy recommendation, which recommends a pedagogical strategy based on the conversation context. Our experiments showed exceptional performance in both classification and recommendation score metrics. Future work will focus on adapting a more diverse and a larger dataset with more pedagogical strategies and scenarios, which would push the model's generalizability further. Additionally, explore a multimodal data inputs such as facial expressions and gesture recognition which would provide a deeper insights and significantly improve Tutor Copilot to emulate a human-like tutor.

% Each one of us should mention what he/she will do in the future in the conclusion part (IMPORTANT!!!!)

\appendix
\section{Additional Experiments}
\label{sec:appendix-experiments}

\subsection{Fine-Grained Classification}

\subsubsection{Additional Experiments}
\label{extra}

Table \ref{tab:model_performancex_resized} compares different models on the augmented balanced dataset.

1) BERT Large Uncased got the highest scores among all different models in validation accuracy 60.27\%, macro F1 score 45.95\%, and test accuracy 56.65\%.

2) Traditional models like decision trees and k-nearest neighbors showed a relatively low performance which reflect their limited ability to handle nuanced and textual relationships.

\bibliography{custom}
\bibliographystyle{acl_natbib}

\end{document}